\documentclass{article}

\usepackage{microtype}
\usepackage{graphicx}
\usepackage{subfigure}
\usepackage{booktabs} 

\usepackage{hyperref}

\usepackage[accepted]{icml2023}

\usepackage{amsmath}
\usepackage{amssymb}
\usepackage{mathtools}
\usepackage{amsthm}

\usepackage[capitalize,noabbrev]{cleveref}

\theoremstyle{plain}

\theoremstyle{definition}

\theoremstyle{remark}

\usepackage[textsize=tiny]{todonotes}
\icmltitlerunning{SAHAAYAK 2023 - the Multi Domain Bilingual Parallel Corpus of Sanskrit to Hindi for Machine Translation}

\begin{document}

\twocolumn[
\icmltitle{SAHAAYAK 2023 - the Multi Domain Bilingual Parallel Corpus of Sanskrit to Hindi for Machine Translation}
\icmlsetsymbol{equal}{*}

\begin{icmlauthorlist}
\icmlauthor{Vishvajitsinh Bakrola}{fa}
\icmlauthor{Dr. Jitendra Nasariwala}{sa}
\end{icmlauthorlist}

\icmlaffiliation{fa}{Assistant Professor, Department of Computer Science and Engineering, Asha M. Tarsadia Institute of Computer Science and Technology, Uka Tarsadia University}
\icmlaffiliation{sa}{Dean, Department of Computer Sciences, Uka Tarsadia University}

\icmlcorrespondingauthor{Vishvajitsinh Bakrola}{vishvajit.bakrola@utu.ac.in}
\icmlcorrespondingauthor{Dr. Jitendra Nasariwala}{jvnasriwala@utu.ac.in}


\vskip 0.3in
]



\printAffiliationsAndNotice{} 

\begin{abstract}
The data article presents the large bilingual parallel corpus of low–resourced language pair Sanskrit – Hindi, named SAHAAYAK 2023. The corpus contains total of 1.5M sentence pairs between Sanskrit and Hindi. To make the universal usability of the corpus and to make it balanced, data from multiple domain has been incorporated into the corpus that includes, News, Daily conversations, Politics, History, Sport, and Ancient Indian Literature. The multifaceted approach has been adapted to make a sizable multi-domain corpus of low-resourced languages like Sanskrit. Our development approach is spanned from creating a small hand-crafted dataset to applying a wide range of mining, cleaning, and verification. We have used the three-fold process of mining: mining from machine-readable sources, mining from non-machine readable sources, and collation from existing corpora sources. Post mining, the dedicated pipeline for normalization, alignment, and corpus cleaning is developed and applied to the corpus to make it ready to use on machine translation algorithms.
\looseness=-1
\end{abstract}

\section{Introduction}
Parallel corpus is a crucial component in the field of machine translation. It refers to a collection of texts in two or more languages, where each sentence in one language has a corresponding sentence in the other language. Building parallel corpora can be a challenging task, especially for low-resource languages. The effectiveness of machine translation systems relies heavily on the quality and size of the corpus used for training. Therefore, building high-quality parallel corpus is essential for achieving accurate and human-like translation results. State-of-the-art Neural machine translation systems are outperforming previous translation approaches. \cite{muskansing} One of the significant limitations of these systems is they are data-hungry. NMT models require a large amount of data to offer good translation accuracy, a particular hindrance from the beginning of the neural machine translation journey.

\section{Need of the Corpus}
Both languages belong to 22 scheduled languages of India under article 343 (1) of the Indian constitution. \cite{article343} In addition, two of the Indian states, i.e., Himachal Pradesh and Uttarakhand, have included Sanskrit as an additional official language, and nine states have Hindi as their first official language. As per the last census of India, 528M native speakers are using Hindi as their mother tongue, and 24.8K native speakers are using Sanskrit as their mother tongue. \cite{census2011} Sanskrit fall under the category of Low Resourced Language (LRL). It becomes a real problem when it comes to the creation of LRLs corpus. These categories of languages are less studied, resource-scarce, less computerized, and less privileged for the time being. \cite{singh-2008-natural} Most of today’s NLP research focuses on 20 of nearly 7000 spoken languages worldwide. This is because of their significantly less digital availability. Sanskrit belongs to the same category in multiple dimensions. Despite its rich grammatical tradition in the linguistic community and being one of the oldest living languages, there was an absolute absence of a general-purpose bilingual corpus with Sanskrit as one of the languages. We cannot consider Sanskrit a resource-scarce and less-studied language, but it is less computerized. Though some good classical Sanskrit texts from Vedas, Ramayana, Mahabharata, and Upanishads are digitized chiefly, they have a monolingual form or are in non-machine-readable formats.

We found that all the previous works contended against the scarcity of sufficient parallel corpus with Sanskrit as one of the languages in the translation pair. We observed that developing a fully-aligned general-purpose parallel corpus keeping Sanskrit as the source language is the first need to establish a state-of-the-art machine translation system. To our knowledge, SAHAAYAK 2023 is the most extensive dataset released in the public domain, with Sanskrit as one of its languages.

\section{General Architecture of Corpus Creation}
The task of corpus creation of LRL required a 360-degree approach for language coverage. To enhance the universal usability the corpus should be balanced. \cite{atkins} We have applied our efforts to balance this corpus by covering various domains. We initiated the parallel corpus development of Sanskrit to Hindi with nearly 3.3K sentences in 2018, and these translations included 2590 human-translated sentences of daily conversations in the Indian context and 701 slokas from the Gita. Over a period, we expanded this further with varieties of mining, human expert verification, and translation by human experts. Our mining approach is divided into three–fold, mining from machine-readable sources, mining from non-machine-readable sources, and Collation from other existing sources of corpora. An overview of our pipeline is shown in Figure – 1.

\begin{figure}[h]
	\begin{center}
		\includegraphics[scale=0.2]{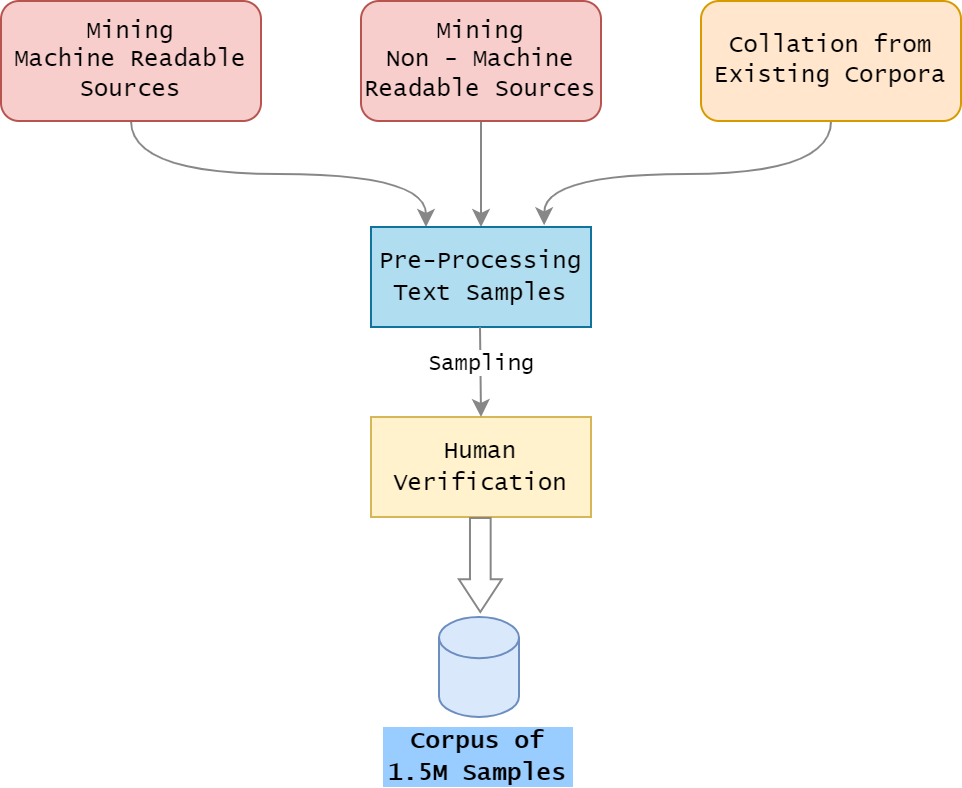}
		\caption[Pipeline of Sanskrit-Hindi Parallel Corpus Creation] {Pipeline of Sanskrit-Hindi Parallel Corpus Creation}
		\label{typical}
	\end{center}
\end{figure}

We identified several web sources which publish content in multiple Indic languages, including Sanskrit. Most of these sources contain comparable texts rather than the direct translation of each other. We have created a source-specific standalone procedure for fetching text data.

For Sanskrit, a considerable amount of physical and scanned documents are available which are not machine-readable. We have developed separate procedures for fetching and aligning text from these sources using OCR. In this mining procedure, there are two varieties of seeds we have worked with. One belongs to those sources which are publicly available data, i.e., scanned copies of Sanskrit and its Hindi translations and physical books, scriptures, and other documents which are in public domains. The dedicated OCR techniques are developed for extraction from such public sources. In addition to these pipelines for parallel text collection, samples were collected from the OPUS, and NLLB. We have also applied parallel text extraction and post-processing steps on these samples to remove the noise and to make the text more clear. Text normalization, Text alignment, and Corpus cleaning techniques are developed and applied keeping in mind the Devanagari script.

\section{Analysis of SAHAAYAK 2023}
Table – 1 provides a glimpse on our bilingual parallel Sanskrit-Hindi corpus.
\begin{table}[ht]
\caption{Analysis of SAHAAYAK 2023}
\label{sample-table}
\begin{center}
\begin{small}
\begin{sc}
\begin{tabular}{lcccr}
\toprule
\textbf{Parameters} & \textbf{Sanskrit} & \textbf{Hindi} \\
\midrule
Number of samples    & 1580701 & 1580701\\
Average Sentence Length    & 64 & 66\\
Longest Sentence Length     & 2005 & 2069\\
\bottomrule
\end{tabular}
\end{sc}
\end{small}
\end{center}
\end{table}

\section{Conclusion}
SAHAAYAK 2023 is an attempt to contribute to the community of machine translation and computational linguistics. We presented a bilingual parallel corpus of 1.5M parallelly aligned sentence pairs from Sanskrit to Hindi. Tens of methods are applied to ensure the overall quality of the corpus in the post–processing. We hope that SAHAAYAK 2023 will be proven helpful in developing state-of-the-art machine translation systems for Indian languages. The improvements in the translation quality of Sanskrit to other Indic languages will also be proven beneficial for Education and different societal needs.

\section{Acknowledgment}
We would like to thank the researchers of NLP and computational linguistic community. We would also like to acknowledge the authors of NLLB and OPUS. We would like to take this opportunity to express our gratitude towards our Sanskrit language experts Shri Harshad Joshi and Shri Udit Sharma for their significant contributions for translation and verification. The acknowledgment is incomplete without showing deep gratitude towards Maharshi Panini, the complete source of inspiration to work in the direction of Sanskrit based computational linguistics and Padmabhusan Shripad Damodar Satwalekar, whose books are proven corner stone for creating interest in the Sanskrit language.

\section{Dataset Source}
The corpus can be accessed at \url{https://rb.gy/hf6bp}
\nocite{langley00}

\bibliography{example_paper}
\bibliographystyle{icml2023}

%

\end{document}